\documentclass[conference]{IEEEtran}
\IEEEoverridecommandlockouts
\usepackage{cite}
\usepackage{amsmath,amssymb,amsfonts}
\usepackage{algorithmic}
\usepackage{graphicx}
\usepackage{textcomp}
\usepackage{xcolor}
\usepackage{float}
\usepackage{balance}
\usepackage[margin=0.75in]{geometry}
\def\BibTeX{{\rm B\kern-.05em{\sc i\kern-.025em b}\kern-.08em
    T\kern-.1667em\lower.7ex\hbox{E}\kern-.125emX}}
\begin{document}

\title{\vspace*{18pt}Force and Speed in a Soft Stewart Platform}

\author{Jake Ketchum$^{1}$, James Avtges$^{1}$, Millicent Schlafly$^{1}$, Helena Young$^{1}$ \\ Taekyoung Kim$^{2}$, Ryan L. Truby$^{1,2}$, Todd D. Murphey$^{1}$ 
\thanks{$^{1}$Department of Mechanical Engineering, Northwestern University, Evanston, IL, USA.
        }%
\thanks{$^{2}$Department of Materials Science and Engineering, Northwestern University,
        Evanston, IL, USA.
        }%
\thanks{This work is supported by the US Army Research Office grant no. W911NF-22-1-0286 and US Office of Naval Research grant no. N00014-21-1-2706). J.A. is supported by a National Defense Science and Engineering Graduate Fellowship. T.K. and R.L.T. acknowledge support from US Office of Naval Research grant no. N00014-22-1-2447 and Leslie and Mac McQuown through Northwestern University’s Center for Engineering Sustainability and Resilience.}
}
\maketitle

\begin{abstract}
Many soft robots struggle to produce dynamic motions with fast, large displacements. We develop a parallel 6 degree-of-freedom (DoF) Stewart-Gough mechanism using Handed Shearing Auxetic (HSA) actuators. By using soft actuators, we are able to use one third as many mechatronic components as a rigid Stewart platform, while retaining a working payload of 2kg and an open-loop bandwidth greater than 16Hz. We show that the platform is capable of both precise tracing and dynamic disturbance rejection when controlling a ball and sliding puck using a Proportional Integral Derivative (PID) controller. We develop a machine-learning-based kinematics model and demonstrate a functional workspace of roughly 10cm in each translation direction and 28 degrees in each orientation. This 6DoF device has many of the characteristics associated with rigid components---power, speed, and total workspace---while capturing the advantages of soft mechanisms. 

\end{abstract}
\color{black}

\section{Introduction}

Soft robots promise to be safer, more resilient, and more adaptable than their rigid counterparts. This is particularly valuable for systems that are expected to touch and interact with people. 6 degree-of-freedom (DoF) positioning is an important robotic capability used in a variety of industries from rehabilitation to flight simulation and industrial shipping. However, existing soft 6 DoF parallel mechanisms struggle to produce the forces, displacements, and response times required for mass adoption \cite{white_soft_2018, kalafat_design_2024, gilmore_development_2024}.

 A substantial driver of this capability gap is the many limitations of soft actuator technologies. Shape memory alloy (SMA) actuators have a long cycle life \cite{disawal_life_2018, reynaerts_design_1998}, produce very large force outputs\cite{reynaerts_design_1998}, and can be quite precise \cite{kim_control_2023}. However, they are also very slow, with even very fast SMA actuators featuring open loop bandwidths of 2Hz or less \cite{russell_improving_1995} and get dangerously hot during operation. By contrast, cable driven soft actuators provide fast, easily-controlled actuation, high force output, and long operating lives \cite{zaidi_actuation_2021}. However, cables are labor-intensive to run and prone to tensioning-related maintenance issues. Many designs also require actuators to be under tension when in their neutral state, creating a high-energy failure mode \cite{abidi_intrinsic_2017}. Pneumatic soft actuators offer high force outputs with large displacements but suffer from limited working lives, catastrophic failure modes, and require a compressor or other source of pressure to operate \cite{zaidi_actuation_2021}. 

Handed shearing auxetic (HSA) actuators are a promising alternative \cite{lipton_handedness_2018}. They feature fast response times with high force outputs and large displacements. HSAs are also responsive, easily 3D printed, and can be controlled via an electronic servo with no additional hardware. In this work, we develop a 6DoF parallel mechanism based on HSAs, which has an open loop bandwidth of 16Hz, a maximum weight capacity of 3.5kg, and an operating life of well over 100 continuous-motion hours. We then demonstrate that this mechanism can be used to complete precise and dynamic manipulation tasks under closed-loop control. We also conduct experiments to quantify its frequency response weight capacity and workspace. 

\begin{figure}
\centering
    \includegraphics[width=0.45\textwidth,keepaspectratio]{"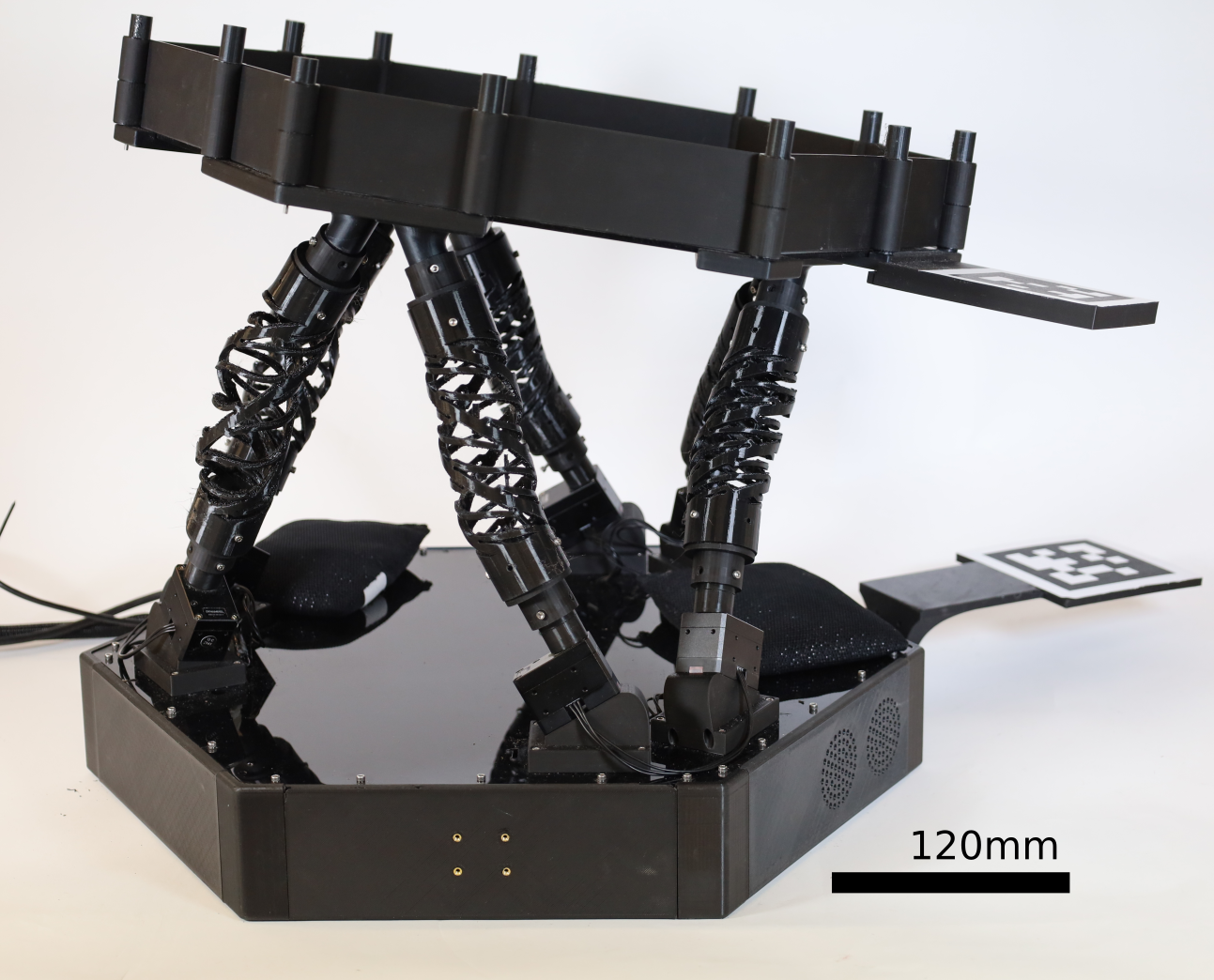"}
\caption{\textbf{Photo and Dimensions of the soft Stewart platform:} 
 The soft Stewart platform at a near-rest position. The platform has a diameter of 40.6cm and a strut-length of 25cm. The platform-to-platform distance ranges from 25.3cm to 33.2cm depending on the system's state. 
}
\label{fig: robot picture}
\end{figure}

The key contributions of this work are: 
\begin{itemize}
    \item  A mechanically simple 6DoF soft parallel mechanism based on HSAs.
    \item Demonstrations of the soft Stewart platform performing precise tracing and dynamic disturbance rejection. 
    \item A frequency response and workspace analysis demonstrating the platform capabilities are competitive with a rigid Stewart platform. 
    \item A data-driven kinematics model and comparison to rigid kinematics. 

\end{itemize}

\section{Related Work}
\label{section: related work}
This work relies heavily on advances in the design of handed shearing auxetic actuators for soft robotics applications. We also make extensive use of prior work in the design and analysis of Stewart-Gough pattern parallel mechanisms. This section provides additional background on both subjects. 

\subsection{Handed Shearing Auxetics (HSAs)}
HSAs are a class of compliant mechanical metamaterial that can act as soft actuators. They are distinguished in this application by their robustness, high force output, and ease of control. Although relatively new, HSAs have been incorporated into a range of soft robotic systems including grippers \cite{chin2018compliant}, quadrupeds \cite{ketchum2023automated,kaarthik2022motorized}, and pipe inspection robots \cite{kim2024flexible}. The HSAs used in this work are 3D printed out of thermoplastic polyurethane (TPU). However, HSAs have also been laser cut out of Teflon \cite{chin2018compliant,chin2020multiplexed, lipton_handedness_2018} and 3D printed from polyurethane photopolymer resins using digital light processing (DLP) \cite{truby_recipe_2021}. 

To actuate an HSA, one connects one end to a servo motor and applies a restoring torque to the opposing end. Rotating the servo-motor causes the HSA to extend or contract, depending on the direction of rotation. To provide the restoring torque, each HSA is typically connected to one or multiple HSAs of opposite handedness. The HSAs are often used in mixed-handedness bundles of 2-4, where each HSA counteracts the shear and torsional forces of the other HSAs \cite{lipton_handedness_2018}. However, HSAs have also been designed with internal baffles that allow them to operate without external constraints \cite{kim2024flexible}. 

Under load, HSAs have nonlinear time-varying dynamics which confound easy simulation. Nevertheless, promising work has been done modeling unloaded 4DoF parallel HSA mechanisms. Piecewise constant curvature models have been used to provide open loop inverse kinematics (IK) solutions to within 5.5mm or 4.5\% of body-length \cite{garg_kinematic_2022}. Tracing tasks have also been demonstrated on 4 HSA bundles using both PID and model-based methods \cite{stolzle_experimental_2024}. However, no work has yet been successful in modeling HSA structures that contain more than 4 elements.

\subsection{Parallel Mechanisms}
The Stewart-Gough Platform, also called the Stewart Platform, is a parallel 6DoF positioning mechanism first developed by V. E. Gough in 1949 to test tires. The design was later publicized by D. Stewart in 1965 as a flight simulator mechanism \cite{stewart_platform_1965}. Since that time, Stewart-Gough platforms have found applications across a wide range of industries ranging from telescope positioning to earthquake tables and gantry stabilization on the high seas. 

The key geometric advantage of a Stewart-Gough mechanism is that it can be attached entirely to one surface, leaving the object under test entirely unencumbered within the mechanism's working range. Commercial Stewart platforms also frequently produce large force and velocity outputs with extremely high precision. However, Stewart-Gough platform's force-production capabilities vary depend on their kinematic configuration making control complicated. Additionally the workspace volume of Stewart platforms can be small relative to their actuator stroke-lengths \cite{dasgupta_stewart_2000}. 

Those same advantages --- workspace, easy mounting, and simplicity --- make the Stewart-Gough mechanism a promising design pattern for soft robotics. 

The existing literature demonstrates that soft parallel mechanisms can feature large displacements and precise motion \cite{huang_kinematic_2022}. Recent work from \cite{gilmore_development_2024} uses hexagonal 3D printed springs compressed by Teflon ligaments to produce a platform with large displacements and fast movement. Similarly, work from \cite{white_soft_2018} uses shape memory alloy actuators to deform a central ball demonstrating that compressive Stewart-Gough platforms can be built without motors. Stewart-like platforms have also been built using unfolding pneumatic actuators \cite{glassner_soft_2020} and origami-inspired living hinges \cite{kalafat_design_2024}. However, there are no known examples in the literature of soft Stewart mechanisms manipulating large loads or performing precise dynamic tasks. These are core capabilities for rigid Stewart platforms without which compliant Stewart-Gough mechanisms will have limited utility. 

\begin{figure*}[h]

\centering
    \includegraphics[width=1\textwidth,keepaspectratio]{"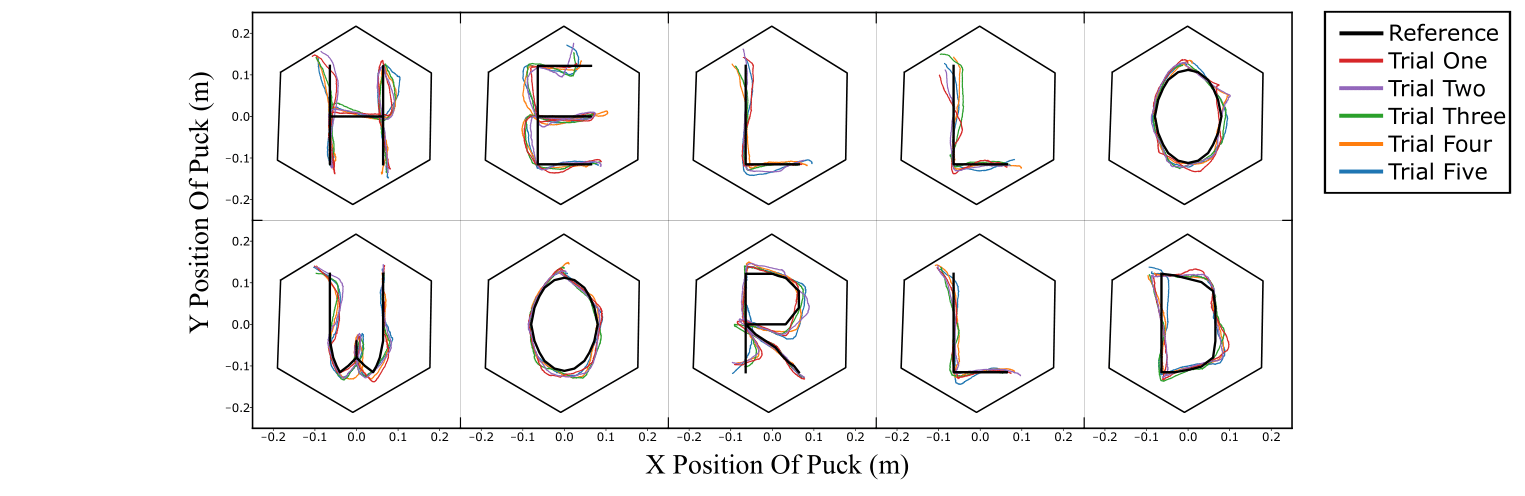"}
    \label{fig pid letters}
\caption{\textbf{Waypoint Letter Tracing:}
 We show that cascading PID control can be used to track waypoints in the shapes of letters spelling ``HELLO WORLD''. Across five trials for each letter, we show that the rolling puck consistently and accurately reaches each waypoint. Letters are traced sequentially, where each letter spans the area of the hexagonal platform.
}
\label{fig: pid letters}
\end{figure*}

\color{black}

\subsection{Soft Robots With Large Displacements and Forces}

Typically, soft robots exhibit large forces by exploiting mechanical instabilities \cite{pal2021exploiting}. 
Although many bistable soft robots require environmental intervention to reverse the phase transition and be capable of reproducing the large force \cite{rojas2022multistable}, some soft robots can repeat the cyclical process without intervention at a particular frequency \cite{kim2021autonomous,comoretto2024physical}. For applications like robot manipulation or agricultural robots, it is important that the soft robot can exhibit the full spectrum of possible forces on command.

To produce large robot displacements, many researchers use pressurized components or pouches containing air or fluid. When differential pressure is applied to different pouches, Kalisky et al.\ show that the robot can bend 79 degrees and apply 2.25N of force \cite{kalisky2017differential}. HASEL actuators can nearly double in size by squeezing a liquid dielectric between two electrodes; however, there is no built-in mechanism for handling out-of-plane forces \cite{mitchell2019easy}. Yang et al.\ use a similar lattice-based structure with pressurized pouches to build a human exosuit, demonstrating the value of large deformations and forces with up to 87.5\% contraction with 20kg of load in separate structures \cite{yang2024high}. In a cyclic loading test, the authors of \cite{yang2024high} report actuation speeds of approximately 0.1m/s.
Despite large displacements, fluidic components can be expensive to manufacture, easily damaged, require a tether, and difficult to control. As an alternative, Mirsa and Sung develop a cable-based soft robot \cite{misra2024online}; however, many components are required to achieve large displacements. In contrast, the soft Stewart platform presented here achieves large displacements with relatively few non-pneumatic components. 

\section{Mechatronic Design}
\label{section: Hardware}

The platform uses six XM430-W350 Dynamixel actuators controlled by a U2D2 USB-to-serial interface module. Internally, the platform uses a Jetson Nano single board computer which handles motor control and the emergency stop system. During these experiments we also use an external computer with a i7-12700k CPU and an RTX 3060 GPU which is connected over USB to an XSense MTI-7 IMU and tracking camera (a Logitech Brio 4k). The external computer is responsible for the computer vision, control code, inverse kinematics solver, and 3D mouse input used for teleoperation. The two computers communicate over Ethernet using ROS2. 

The camera runs with a resolution of 1280 x 720 and 90Hz and is used to track AprilTags on the top and bottom of the platform. We also use it to track a ball for our PID experiments, and apply a 4 sample moving average prior to publishing its pose. After processing, the ball pose is published at a rate of 45Hz. The motors publish current, velocity, position, and temperature information at 30Hz, and can process commands at a maximum of 100Hz.

We use a Stewart platform pattern for this robot as shown in Figure \ref{fig: robot picture}. The bottom end of each strut is set on a 175mm diameter circle, and the top of each strut set on a 152mm diameter circle, producing a 10 degree inward slant. The servo motors and HSA tops are set 15.5 degrees from their associated corners, and each strut is 250mm long including the HSA and coupler assemblies. The body of the robot is made from a combination of laser cut acrylic and 3D printed polylactic acid (PLA). 

The platform struts are 3D printed TPU HSAs similar to the type described in \cite{kim2024flexible}. The HSAs have an inner diameter of 33.5mm and a wall thickness of 4mm. We had originally used HSAs with an ID of 24.5mm and a thickness of 3mm but found that they had a tendency to buckle when fully extended. The HSAs have a repeat unit length of 80mm and a maximum extension of 50mm. During operation, these HSAs require 0.4Nm, about 10\% of the motors' maximum, and we did not see any significant asymmetry in extension and contraction speeds.  
    
One key advantage of soft robots over their rigid counterparts is reduced part count. As outlined in Table \ref{tbl: mechanical elements}, a rigid Stewart platform typically requires either five or six mechatronic elements per strut excluding fasteners, for a total of around 30 per platform. However, because soft actuators like the HSAs used in this work can provide structure, actuation, and compliance in a single component, the soft Stewart platform showcased in this work requires only two mechatronic elements per strut for a total of 12. \

\color{black}
\begin{table}
\vspace{-10pt}
\begin{center}
\caption{Mechanical Complexity of Stewart Platform Strut Designs}
\begin{tabular}{ |p{2cm}|p{0.9cm}|p{1.2cm}|p{1cm}|p{1.5cm}|  }
\hline
 Actuator Type & Servo Motors & Motion Components & Linkages & Total Elements Per Strut \\ 
 \hline
 Soft Stewart    & 1 & \multicolumn{2}{c|}{1 (HSA)} & 2  \\  
 \hline
 Linear Actuator & 1 & 2 & 2  & 5   \\ 
  \hline
 Rotary Arm      & 1 & 4 & 1 & 6   \\
  \hline
\end{tabular}

\label{tbl: mechanical elements}
\end{center}
\vspace{-10pt}
\end{table}

\subsection{Stewart Kinematics}
\label{subsec: Stewart Kinematics}
We found that an appropriately calibrated rigid kinematics approximation of the soft Stewart platform performed well for moderate loading cases. This is valuable because the inverse kinematics model for a Stewart parallel linkage has a trivial closed-form solution \cite{murray_mathematical_1994}. This makes the rigid approximation perfect for gross high frequency control of the kind required for manipulating objects on the platform's surface using PID, and for teleoperation using a 3Dconnexion SpaceMouse. To accurately model the platform's nonlinearities we also train a neural network based learned kinematics model which is described in Section \ref{section: method}.

 The inverse kinematics model for a rigid Stewart platform can be defined by four values: 
\begin{itemize}
    \item $R_u$ describing the radius on which the struts are set on the upper platform. 
    \item $R_l$ describing the radius on which the struts are set on the lower platform. 
    \item $\theta_o$ describing the magnitude of the angle from each strut connection to its associated corner.
    \item $x$ the 6DoF pose of the top platform in the frame of the bottom platform. 
\end{itemize}

Let us now begin by specifying that the function $T_{A\_B}(x, y, z, \theta, \phi, \phi)$ will represent the homogeneous transformation matrix associated with the transform from some frame $A$ to frame $B$ as shown on page 76 of \cite{Lynch_Park_2017}.

Now let us define two frames $U$ and $L$ representing the center of the platform's upper and lower surfaces respectively. We can then further define frames $SLn$ and $SUn$ for $n \in [1..6]$ which sit at the upper and lower attachment points of each strut. The locations of the upper and lower ends of each strut are then given by (\ref{equ: strut bottoms}) and (\ref{equ: strut tops}). 

\begin{equation}
\theta_n = sign(n) \theta_o + \frac{2}{3}\pi \left \lfloor{\frac{n}{2}}\right \rfloor 
\label{equ: theta n}
\end{equation}

\begin{equation}
T_{L\_SLn} = T(\cos(\theta_n)R_l, \sin(\theta_n)R_l, 0, 0, 0, 0)
\label{equ: strut bottoms}
\end{equation}

\begin{equation}
T_{U\_SUn} = T(\cos(\theta_n)R_l, \sin(\theta_n)R_l, 0, 0, 0, 0)
\label{equ: strut tops}
\end{equation}

The vector describing each strut is then given by (\ref{equ: strut vect}) and the length of each strut $l_n$ by (\ref{equ: strut length}). Since the extension of an unloaded HSA is approximately linear with its rotation, the appropriate servo angles can be calculated directly from these strut length values. 

\begin{equation}
T_{SLn\_SUn} = T^{-1}_{L\_SLn} T_{L\_U}T_{U\_SUn}
\label{equ: strut transform}
\end{equation}

\begin{equation}
\vec{Sn} = T_{SLn\_SUn} \begin{bmatrix}
           0 \\
           0 \\
           0 \\
           1 \\
         \end{bmatrix}
\label{equ: strut vect}
\end{equation}

\begin{equation}
l_n ={ \|\| \vec{Sn}\|\|}
\label{equ: strut length}
\end{equation}

\color{black}

\section{Method}
\label{section: method}

To evaluate our proposed mechanism, we conducted a variety of experiments meant to explore the soft Stewart platform's operating capabilities. The frequency response analysis, workspace characterization, and learned kinematics model provide a quantitative benchmark of the platform's capabilities. The PID tracing and disturbance rejection tasks are both analogs to common demonstrations performed by rigid Stewart platforms \cite{PPOD_VOSE2013111} and are meant to provide a qualitative sense of the system's capabilities. Additional details about how each experiment was conducted can be found in the following sub-sections. 

Three different sets of HSAs are used for this work. The first set started with around 120 operating hours and is used for the PID figure. The second set was freshly printed and is used for load capacity testing, frequency sweeps, workspace analysis, and learned kinematics. A third set was also freshly printed for this work and is used exclusively in the supplemental video. While we have never had an HSA become inoperable while on the platform, we have noticed that the platform dynamics change as the HSAs wear. This behavior is already noted in \cite{kim2024flexible, truby_recipe_2021} and is documented there in more detail. 

\color{black}
\subsection{PID Control}
We implement a cascading PID controller for the platform, which enables detailed point-to-point tracing and dynamic disturbance rejection tasks. The separation of the PID control into two discrete loops iterating over velocity and position allows for precise tuning and quick response times.

\begin{figure}[h]

\centering
    \includegraphics[width=0.45\textwidth,keepaspectratio]{"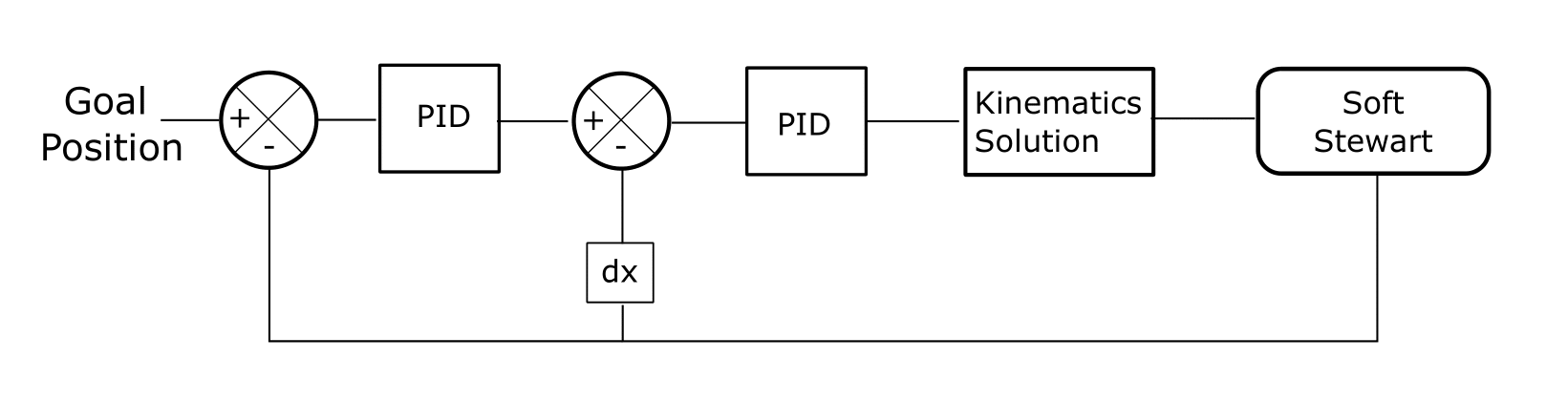"}\caption{\textbf{System Control Diagram: }
 The control architecture used for our manipulation experiments. The inner PID loop handles ball velocity and the outer control loop handles ball position. $dx$ represents the velocity term used to compute error for the inner loop. Each axis $x$ and $y$ has a separate cascade controller. 
}
\label{fig: control diagram}
\end{figure}

The outer PID loop controls for the object's position. It takes a desired waypoint reference, calculates position error in $x$ and $y$ in the platform surface frame $U$ from camera data, and uses the PID controller's output as the reference ``desired velocity'' for the inner velocity controller. The inner PID loop uses the difference between the measured and desired velocity in $x$ and $y$ as its error term, and the control output in desired roll and pitch of the platform is fed into the rigid kinematics model as depicted in Figure \ref{fig: control diagram}, determining motor angles. In the waypoint following task, the next waypoint in the path is incremented every eight seconds.

For the dynamic disturbance rejection task, we use a smooth ball; for the precision tracking task a plastic ring is placed over the ball to introduce static friction (referred to as a ``puck''), though the ball inside is still allowed to roll. Both weigh approximately 50 grams.

\begin{figure*}[ht]

\centering
    \includegraphics[width=0.90\textwidth,keepaspectratio]{"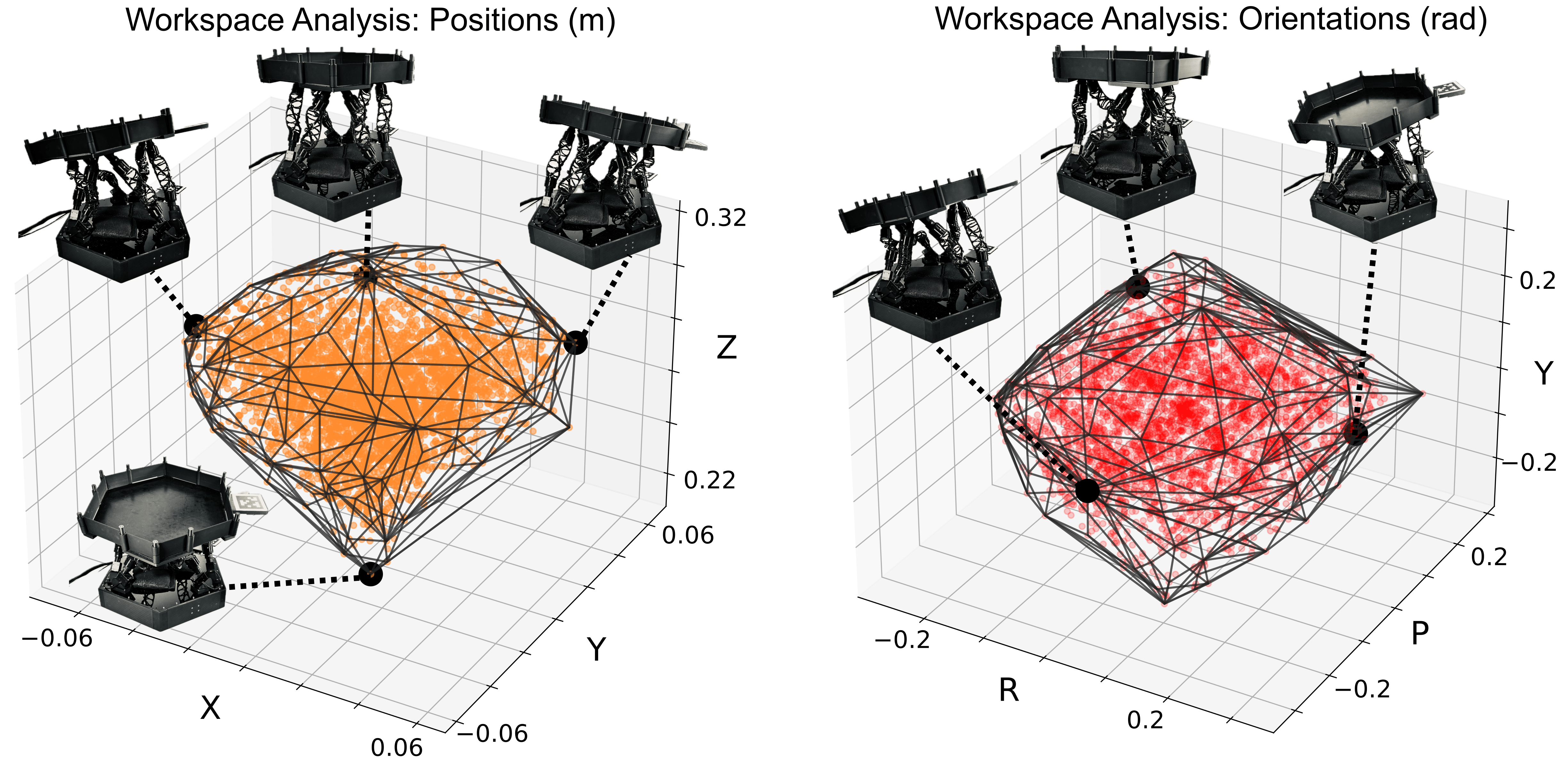"}
\caption{\textbf{Mechanism Workspace:  }
We performed a raster scan with 90-degree resolution over all combinations of motor angles from $[0, 270]$ degrees. The resulting positions and orientations are displayed with a convex hull for clarity. We found the X-Y cross-section of the position data to be approximately triangular with the extreme points (shown) opposite the corners where two servo motors are mounted. The maximum Z position is also displayed. For orientation, we display the poses in the workspace with maximum roll, minimum pitch, and maximum yaw.}
\label{fig: workspace}
\end{figure*}

\subsection{Frequency Analysis}

For each of roll, pitch, and yaw, we input a series of 30 fixed-amplitude sinusoids spanning a range from 0.1Hz to 30Hz with constant spacing in log-space. The resulting platform motions were then recorded using an XSense MTI-7 IMU placed below the center of the platform. We set the amplitude to 10\% of each axis's working range and applied an additional 20\% positive offset in Z to allow for smooth platform motion. Platform positions were recorded from the IMU's onboard filter, and both commands and system states were recorded at 100Hz. ROS2, which we used for this test, introduces its own latency. By placing the IMU on a motor during operation we determined that the combined command and measurement latency to the motor and from the IMU is approximately 8ms. Each input frequency was tested for a period of 60s with a 3s delay between frequencies to allow the platform to reset. 

To process the output amplitude, we compute a one-wavelength rolling average minimum and maximum value for the signal. We then took the mean of this value, normalized it against the 0.1Hz amplitude, and plotted it on a Decibel scale, which can be found in Figure \ref{fig: bode plot}. We originally used a band-pass filter to remove low-frequency drift, but an examination of the underlying data showed no evidence of drift over the 60s collection periods. To produce a phase-delay value, we first shifted the input and output signals appropriately to remove the 8ms of command and measurement delay introduced by ROS2 and our local LAN network. We then found the maximum correlation delay time and computed a corresponding phase delay as shown in Figure \ref{fig: bode plot}.

\color{black}

\subsection{Workspace Analysis}
To analyze the workspace of the platform, we perform a shuffled raster scan over all joint angle combinations with 90-degree increments between [0, 270] degrees. This produces a total of 4096 different poses. Each pose is held for 3s each before recording state information, and the platform returns to its neutral state after each pose. Figure \ref{fig: workspace} shows the point clouds corresponding to position and orientation produced by this analysis along with sample poses at the bounds of the workspace. We find that the point cloud for position has an approximately triangular shape --- the vertices of which match with the opposites of where the HSA struts meet at the top of the platform. This workspace sweep was performed on new but broken-in HSAs with approximately 3 hours of prior usage. We report the approximate workspace bounds in Table \ref{tbl: workspace limits}.

\subsection{Learned Kinematics}
\vspace{-5pt}
 As described in \ref{subsec: Stewart Kinematics}, a rigid approximation of the platform's inverse kinematics allows for imprecise control of the platform tilt, suitable for dynamic balancing of a rolling ball or sliding puck. However, this approximation is impractical for more precise positioning tasks and displacements that require larger extensions of the HSAs. To better capture the nonlinearities of the platform, we trained a neural network to learn the inverse kinematic function, using a training set sampled from static platform poses collected during the workspace analysis along with an additional 5904 poses sampled randomly from the control space. The neural network architecture is three fully-connected layers with rectified linear unit (ReLU) activation and 128 neurons each. The input is six-dimensional, as we represent the platform orientation using Euler angles. The output is also six-dimensional, with the joint angles in degrees to achieve the desired input pose.

\begin{figure}[ht]

\centering
    \includegraphics[width=0.45\textwidth,keepaspectratio]{"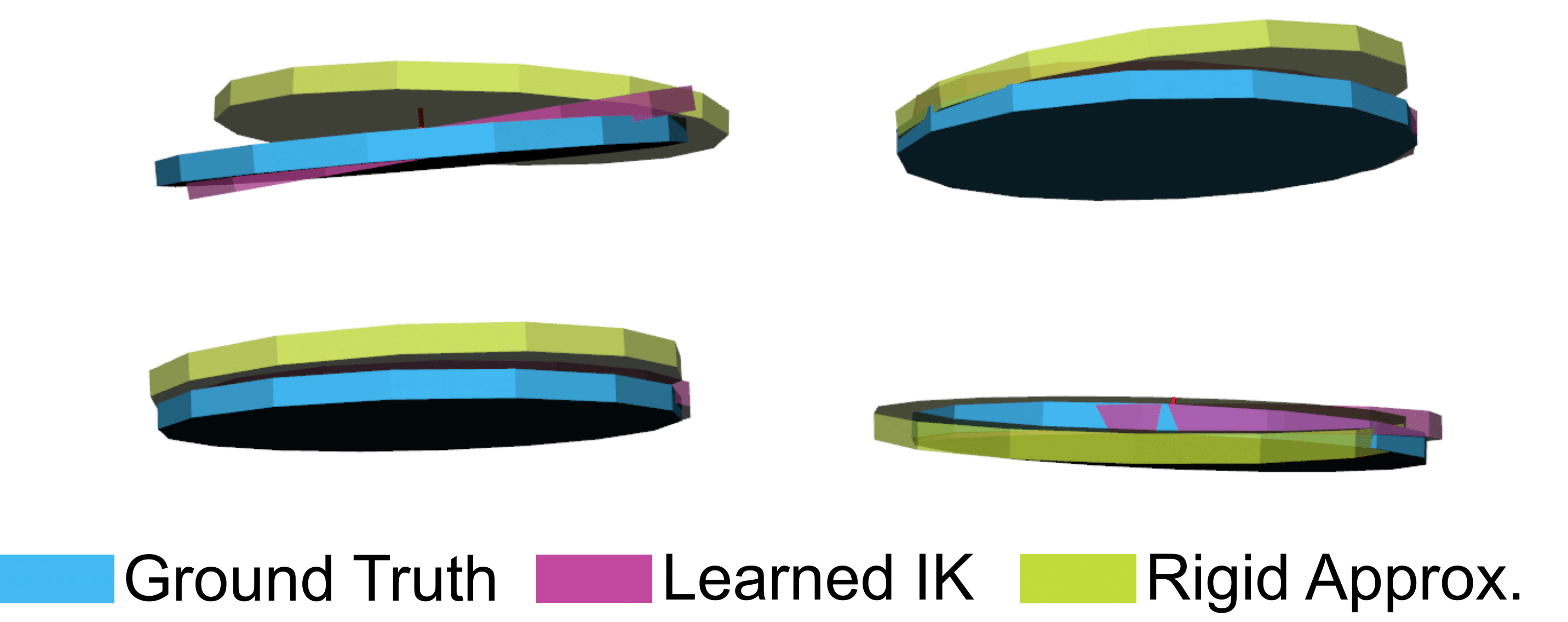"}
\caption{\textbf{Learned Kinematics Model Comparison:}
We compare sample platform poses produced by our learned inverse kinematics model (purple) to a target pose (blue) and a rigid approximation (green). We tested 100 randomly generated poses which are not in the training set and found that in each dimension the learned IK model is accurate to within 4.0mm and 1.15 degrees on average. 
}
\label{fig: ik}
\end{figure}

\section{Results}
\label{section: results}

\begin{figure}[ht]

\centering
    \includegraphics[width=0.5\textwidth,keepaspectratio]{"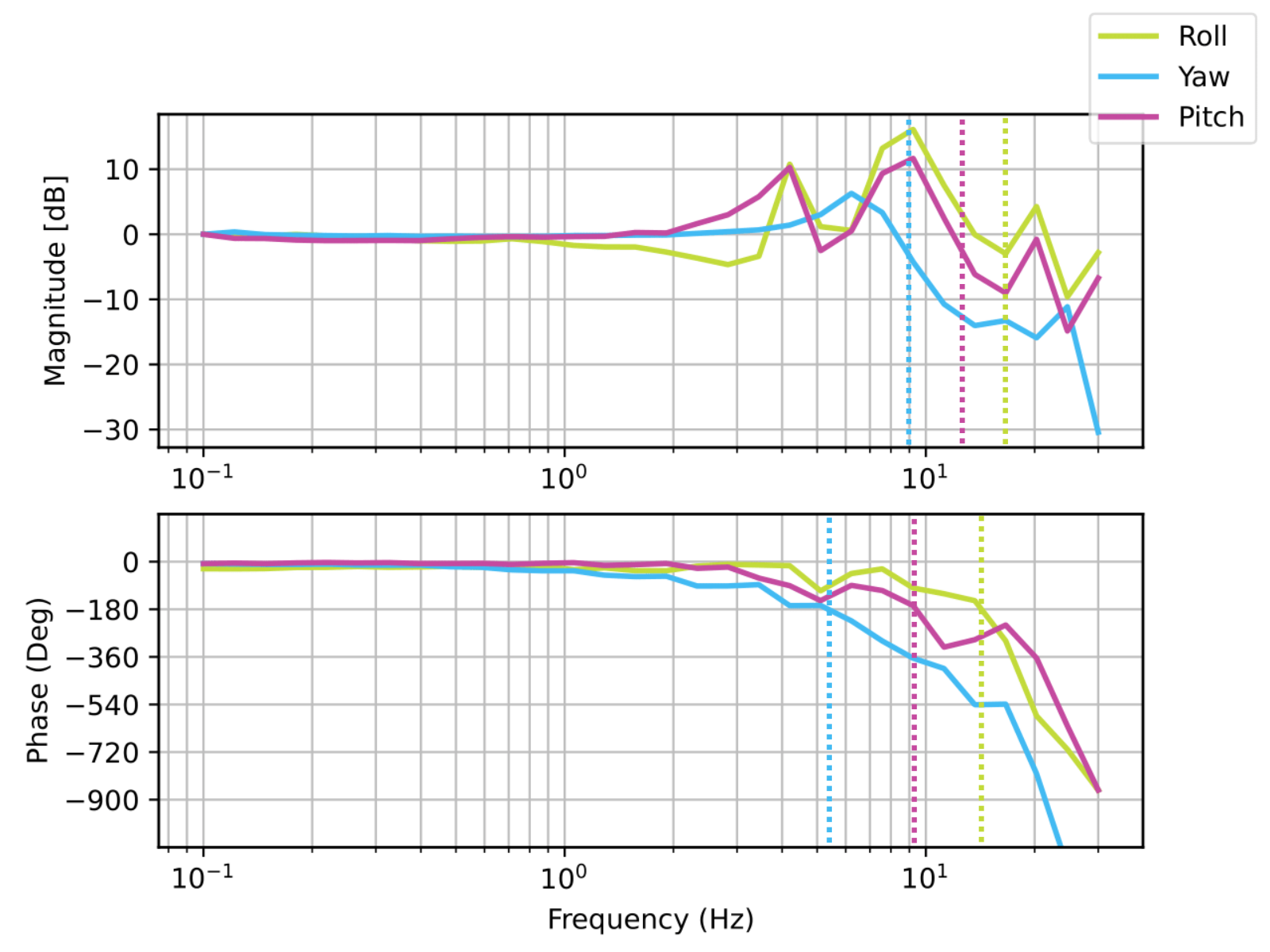"}
\caption{\textbf{Frequency Response Plots:}
This Bode plot shows the frequency response of the robot in roll, pitch, and yaw for frequencies between 0.1 and 30Hz. We note resonant peaks at 4Hz, 6Hz, 9Hz, and 20Hz, which can be excited during closed-loop control. These resonant peaks should be considered when choosing a control frequency for this system. The vertical bars show the frequency at which each frequency sweep crosses the -3db and -180 degree lines for the magnitude and phase plots respectively.}
\label{fig: bode plot}
\end{figure}

\subsection{Task Performance}
Our goal with this robot was not to replicate the performance of rigid Stewart platforms: soft actuators bring their own challenges and advantages, but we do want to show that our mechanism can perform some of the same tasks commonly used to demo conventional Stewart-Gough mechanisms. Figure \ref{fig pid letters} shows the platform's performance for a high-precision tracing task. For this experiment, we use the platform to navigate a series of waypoints with the puck to spell out the letters in ``Hello World". Each letter in ``Hello World" is 22cm high, has between 24 and 44 waypoints, and took between 2 and 7 minutes to complete depending on the path-length of the letter. We repeated each trajectory five times for a total of 50 trajectories. As shown in Figure \ref{fig: pid letters}, all fifty trajectories are easily distinguishable as their respective letters and demonstrate high accuracy, with a Mean Squared Error (MSE) across all trials of 0.085cm\textsuperscript{2}. While this trial was completed using worn-in actuators (see section \ref{section: method}), we repeated the same letter trajectories using brand-new actuators and found no difference in platform performance (see supplemental video). 

 By contrast, Figure \ref{fig: disturbance} shows a low-precision, but highly dynamic task: disturbance rejection. For this task, we selected a 1.5in Delrin ball with a very low coefficient of rolling friction to minimize the system's passive damping. The ball was allowed to stabilize for at least 10s  and was then perturbed by hand with a meter stick. This process was repeated three times for three different directions of disturbance, and the results are complied into Figure \ref{fig: disturbance}. The control signal noise seen in column 3 of Figure \ref{fig: disturbance} comes from ball tracking noise and the derivative term of our PID controller. We found that on average, the platform rejected 95\% of each disturbance within 20s. This demonstrates the platform's ability to perform dynamic tasks under closed-loop control, even in the presence of imperfect sensing. Additional demonstrations of disturbance rejection are shown in the supplementary video.
\begin{figure}[ht]

\centering
    \includegraphics[width=0.5\textwidth,keepaspectratio]{"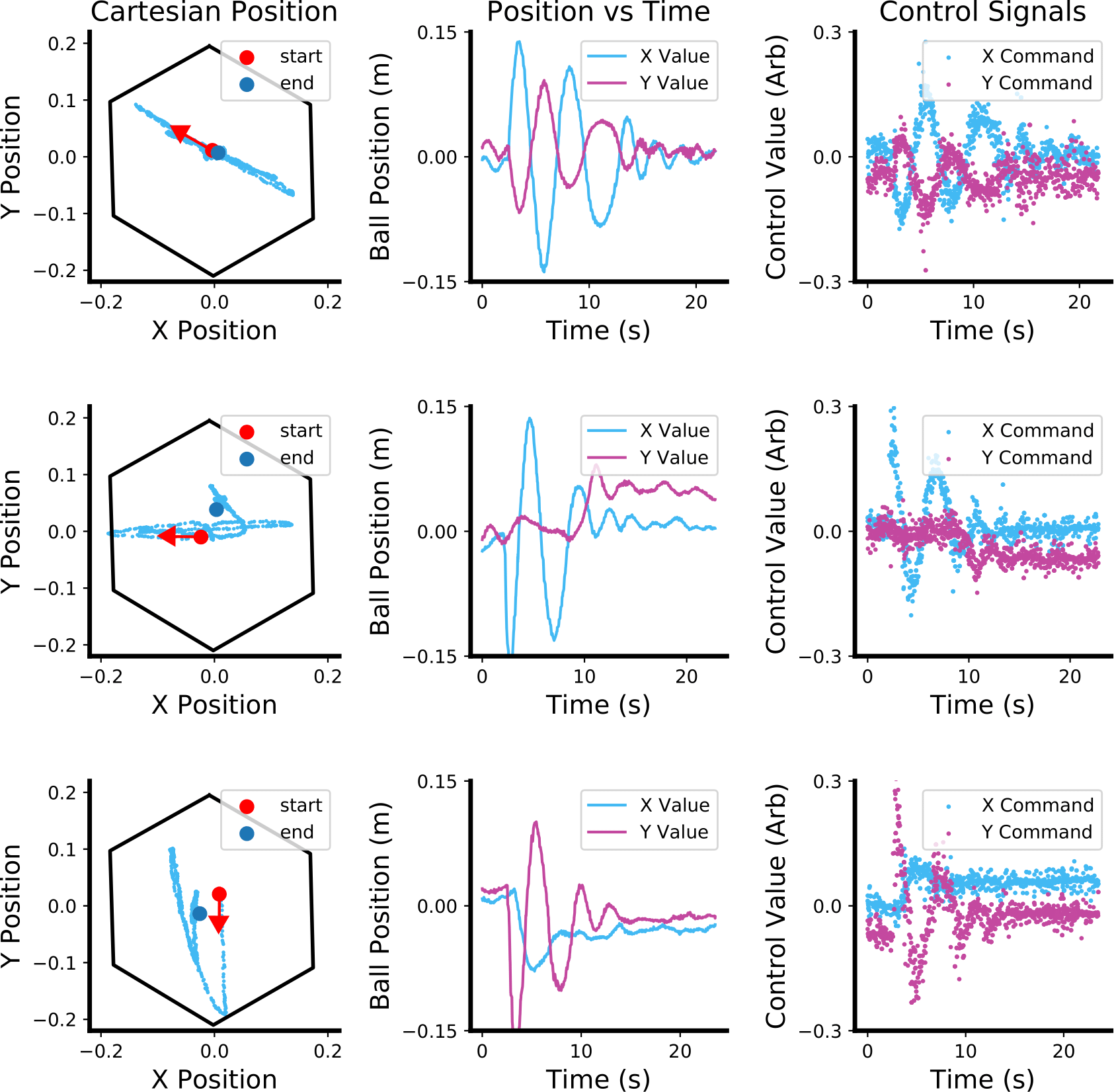"}
\caption{\textbf{Disturbance Rejection: }
 Tests of the platform performing disturbance rejection with a low-friction 1.5in Delrin ball. The ball was allowed to stabilize in the center of the platform before being pushed out with a stick. Each row corresponds to a different different trial. The first column shows the trajectory of the ball in space, the second column shows the trajectory of the ball in time, and the third column shows the control response of the platform. A red arrow in column one shows the direction of impulse disturbance applied to the ball. Note that while there is some persistent offset in trial 2, the platform successfully stabilizes the ball within 20s in all three cases. 
}
\label{fig: disturbance}
\end{figure}

We qualitatively found that the platform's passive rebound to be both substantial and variable depending on z-height of the platform. This configurable passive compliance is a property that conventional Stewart platforms do not share and can only be replicated using powerful motors run under high-frequency closed-loop control. This suggests a range of applications requiring moderately accurate positioning with compliance, for example where human contact is expected, for which soft 6DoF mechanisms may ultimately prove both cheaper and more suitable.

\subsection{System Characterization}

\begin{table}
\vspace{-10pt}
\begin{center}
\caption{Inverse Kinematics Model Comparison}
\begin{tabular}{ |p{1.0cm}|p{1.2cm}|p{1.6cm}|p{1.4cm}|p{1.1cm}|  }
\hline
Axis & Learned Model & Rigid \newline Approximation & Improvement & Units\\ 
 \hline
 X & 0.004 & 0.013 & 72\% & m  \\  
  \hline
   Y & 0.003 & 0.015 & 78\% & m  \\  
  \hline
   Z & 0.005 & 0.017 & 69\% & m  \\  
  \hline
   Roll  & 0.69 & 3.58 & 80\% & Deg  \\  
  \hline
   Pitch  & 1.37 & 2.86 & 52\% & Deg  \\  
  \hline
   Yaw   & 1.37 & 2.55 & 46\% & Deg  \\  
  \hline

\end{tabular}

\label{tbl: ik error}
\end{center}
\vspace{-10pt}

\end{table}

In this subsection we discuss the platform's responsiveness, weight capacity, and reachable workspace. 

To test the platform's weight capacity, we loaded the platform with progressively larger weights until it began to experience buckling in some parts of the workspace. We found that the platform has a maximum load of 3.5kg, defined as the largest weight which allows half the control space to be used, and a working load of 2kg, which allows the full control space to be used. These load values are in addition to the 1kg weight of the platform and fencing, which could be modified or reduced to allow for a working load of 3kg with a maximum load of 4.5kg.

The results of our workspace characterization experiments can be found in Figure \ref{fig: workspace} and are summarized in Tables \ref{tbl: correlation} and \ref{tbl: workspace limits}. We found that the platform has a linear working range of 8cm in $z$ and at least 11cm in $x$ and $y$ with at least 30 degrees of range in roll, pitch, and yaw. The strut configuration we selected --- which features a relatively shallow tilt angle --- exhibits a high correlation between tilts and horizontal movements as shown in Table \ref{tbl: correlation}, but a relatively lower coupling between rotation and z motions. This balance could be adjusted as needed by changing the strut angle and associated strut mounts.

As configured, the platform has a maximum unloaded speed of 0.1m/s and a maximum loaded (2kg) speed of 0.09m/s. The unloaded frequency response of the system can be found in Figure \ref{fig: bode plot} for all three rotational DoF of the mechanism. As shown, the system has an open-loop roll bandwidth of 16Hz and reaches a phase delay of -180 degrees at 14Hz, which is more than sufficient for closed-loop control of both the platform itself or a rolling/sliding object on the platform's surface.

It is interesting to note that pitch exhibits a slightly slower frequency response than roll. We believe this is related to the alignment of the axes relative to the 3 pairs of HSA struts on the platform. We did not observe this effect during task-performance testing but did observe resonant behavior consistent with the two lower peaks at 4Hz and 9Hz shown in Figure \ref{fig: bode plot}. In practice, we were able to manage the systems' resonant behavior by filtering our perception pipeline to avoid injecting noise.

\subsection{Inverse Kinematics}

We experimented with two forward kinematics solutions in this work. Our rigid kinematics model is based on the classic Stewart mechanism kinematics described in section \ref{section: method} and makes no accommodations for the compliant nature of the mechanism. The advantage of this model is that it can be trivially defined using the platform's geometry and introduces essentially no computational overhead. We used it for both manipulation tasks and found it to be more than adequate for that purpose. Its average error across the workspace is 14.8mm in $x$, $y$, and $z$ and 3.00 degrees in roll, pitch, and yaw for each dimension. Its angular accuracy is well-suited for balancing tasks, however, its lack of precise positioning may limit the model's utility for other tasks.

\begin{table}
\caption{Workspace Correlation}
\begin{center}
\begin{tabular}{|p{0.8cm}|p{0.8cm}|p{0.8cm}|p{0.8cm}|p{0.8cm}|p{0.8cm}|p{0.8cm}|}

\hline

      & X     & Y    & Z     & Roll  & Pitch & Yaw   \\ \hline

X     & 1     & -0.00 & -0.12 & 0.04  & 0.78  & -0.01 \\ \hline

Y     & -0.00 & 1     & 0.03  & -0.70 & 0.00  & 0.04  \\  \hline

Z     & -0.12 & 0.03  & 1     & 0.02  & -0.20 & -0.08 \\  \hline

Roll  & 0.04  & -0.70 & 0.02  & 1     & 0.03  & 0.00  \\ \hline

Pitch & 0.78  & 0.00  & -0.20 & 0.03  & 1     & 0.01  \\ \hline

Yaw   & -0.01 & 0.04  & -0.08 & 0.00  & 0.01  & 1    \\ \hline

\end{tabular}
\end{center}
\label{tbl: correlation}
\end{table}

Our learned kinematics model is trained on 10k data points and introduces additional compute and logistics constraints, but it is substantially more accurate. Compared to our rigid model, the learned IK model is 73\% more accurate in translation and 62\% more accurate in rotation. It has an average linear and angular error of 4.0mm and 1.15 degrees respectively. More detailed information can be found in Table \ref{tbl: ik error}.
Figure \ref{fig: ik} shows a comparison between the two models for various target positions. Although the learned model is clearly more accurate, the rigid kinematics model may be sufficient for low-precision tasks, or for moderate-precision tasks when paired with a closed-loop controller. 
\balance

\begin{table}
\vspace{-10pt}
\begin{center}
\caption{Workspace Limits}
\begin{tabular}{ |p{1.2cm}|p{1.5cm}|p{1.5cm}|p{1cm}|p{1.3cm}|  }
\hline
 Axis & Minimum & Maximum & Total & Units\\ 
 \hline
 X & -0.06 & 0.06 & 0.12 & m  \\  
  \hline
   Y & -0.056 & 0.061 & 0.117 & m  \\  
  \hline
   Z & 0.237 & 0.316 & 0.079 & m  \\  
  \hline
   Roll & -15.5 & 18.6 & 34.0 & Degrees  \\  
  \hline
   Pitch & -18.2 & 16.0 & 34.3 & Degrees  \\  
  \hline
   Yaw   & -14.8 & 14.4 & 29.7 & Degrees  \\  
  \hline

\end{tabular}

\label{tbl: workspace limits}
\end{center}
\vspace{-10pt}

\end{table}

\color{black}

\section{Conclusion}

Here we present a compliant Stewart-Gough mechanism based on 3D printed HSAs. Our mechanism features a 3.5kg maximum load, with a 12cm range of motion, and has an operating life in excess of 120 movement-hours. Using a learned controller the platform can be positioned to within 4.0mm and 1.15 degrees. With these capabilities we show that the platform is capable of performing standard capability demos for rigid Stewart-Gough platforms including a dynamic disturbance rejection task and a precise puck-tracing task. 

The development of soft robotic positioners with practical force, displacement, and bandwidth characteristics has the potential to be transformational for robots operating in delicate or human-rich environments. In future work, we plan to leverage the platform's large force output and responsive motion for real-world tasks including digging, mechanical assembly, and tactile perception. We also plan to experiment with the use of integrated soft-sensors to enable high-fidelity force-torque sensing with no meaningful increase in mechanical complexity \cite{truby2022fluidic}. Since operating life and susceptibility to wear are problems inherent to the use of soft actuators, we plan to explore learning-based methods for detecting, analyzing, and adapting to HSA damage in real-time. This will help enable the practical deployment of HSA-based soft positioners in unstructured environments over protracted periods of time.

\clearpage

\bibliographystyle{ieeetr}

\bibliography{soft_stewart}

\balance

\end{document}